\documentclass{article}
\RequirePackage{subfigure}
\usepackage{spconf,amsmath,graphicx}
\usepackage{hyperref}
\usepackage{multicol}
\graphicspath{ {images/} }
\title{Joint Object and State Recognition using Language Knowledge}
\name{Ahmad~Babaeian~Jelodar,~and~Yu~Sun
\thanks{}}
\address{Department of Computer Science and Engineering, University of South Florida, Tampa, FL, USA}
\begin{document}
%\ninept
%

\onecolumn

\begin{flushleft}
\end{flushleft}

\Huge
\noindent
\begin{flushleft}
IEEE Copyright Notice
\end{flushleft}

\normalsize
\noindent
\textcopyright \enspace 2019 IEEE. Personal use of this material is permitted. Permission from IEEE must be obtained for all other uses, in any current or future media, including reprinting/republishing this material for advertising
or promotional purposes, creating new collective works, for resale or redistribution to servers or lists, or
reuse of any copyrighted component of this work in other works.
\linebreak
\linebreak
\linebreak

\Large
\noindent
Accepted to be Published in: Proceedings of the 2019 IEEE International Conference of Image Processing (ICIP), September 22-25, 2019, Taipei, Taiwan.

%\newpage\null\thispagestyle{empty}\newpage

\twocolumn
\normalsize
\maketitle

\begin{abstract}
The state of an object is an important piece of knowledge in robotics applications. 
States and objects are intertwined together, meaning that object information can help recognize the state of an image and vice versa. 
This paper addresses the state identification problem in cooking related images and uses state and object predictions together to improve the classification accuracy of objects and their states from a single image. 
The pipeline presented in this paper includes a CNN with a double classification layer and the Concept-Net language knowledge graph on top. 
The language knowledge creates a semantic likelihood between objects and states.
The resulting object and state confidences from the deep architecture are used together with object and state relatedness estimates from a language knowledge graph to produce marginal probabilities for objects and states. 
The marginal probabilities and confidences of objects (or states) are fused together to improve the final object (or state) classification results. 
Experiments on a dataset of cooking objects show that using a language knowledge graph on top of a deep neural network effectively enhances object and state classification.

\end{abstract}
\section{Introduction}
\label{sec:intro}

Image classification is a research area in computer vision that has gained great attention in recent years mainly to tackle object classification and detection problems \cite{Resnet,Review_ActionRecognition1,Review_ImageCaption1}. 
Object states, on the contrary, have not been considered as much as object classification in recent literature.
Moreover, object states require further analysis especially for robotics-based applications. 
Robotic manipulation, task planning, and grasping require knowledge and constant feedback about the state of the environment and objects.
For instance, if a robot chef wants to perform the task of chopping an onion, it has to grasp the whole onion, cut it into half, recognize its new state (sliced), grasp it accordingly, and cut it into smaller parts while continuously monitoring the state.
Ultimately, the robot needs to recognize the desired state and understand when it has reached the end of the procedure (e.g. chopping).
The problem of states has been analyzed in several previous works \cite{added4_objectstates,added1_multi_task_cnn,Ahmad_Paper}.
Similar to \cite{Ahmad_Paper} we will address the issue of states in cooking related images.

States of objects are not independent of the object itself, the action happening, or the scene. Additional information from a single image such as knowledge about the objects in the image will lead to more accurate state classification results. 
Some research has focused on joint state and action or state and object classification \cite{Review_ActionRecognition1}.
Language knowledge graphs are useful for analyzing semantic relationships \cite{concept_net1,Word2Vec}. 
Language knowledge graphs can draw a connection between objects and states in an image and define the likelihood of an object and state occurring together.
Combining the image classification power of deep convolutional networks with the semantic power of a language knowledge can provide a powerful tool for joint state and object classification. 

\begin{figure*} [!h]
   \centering
   \includegraphics[width=16cm]{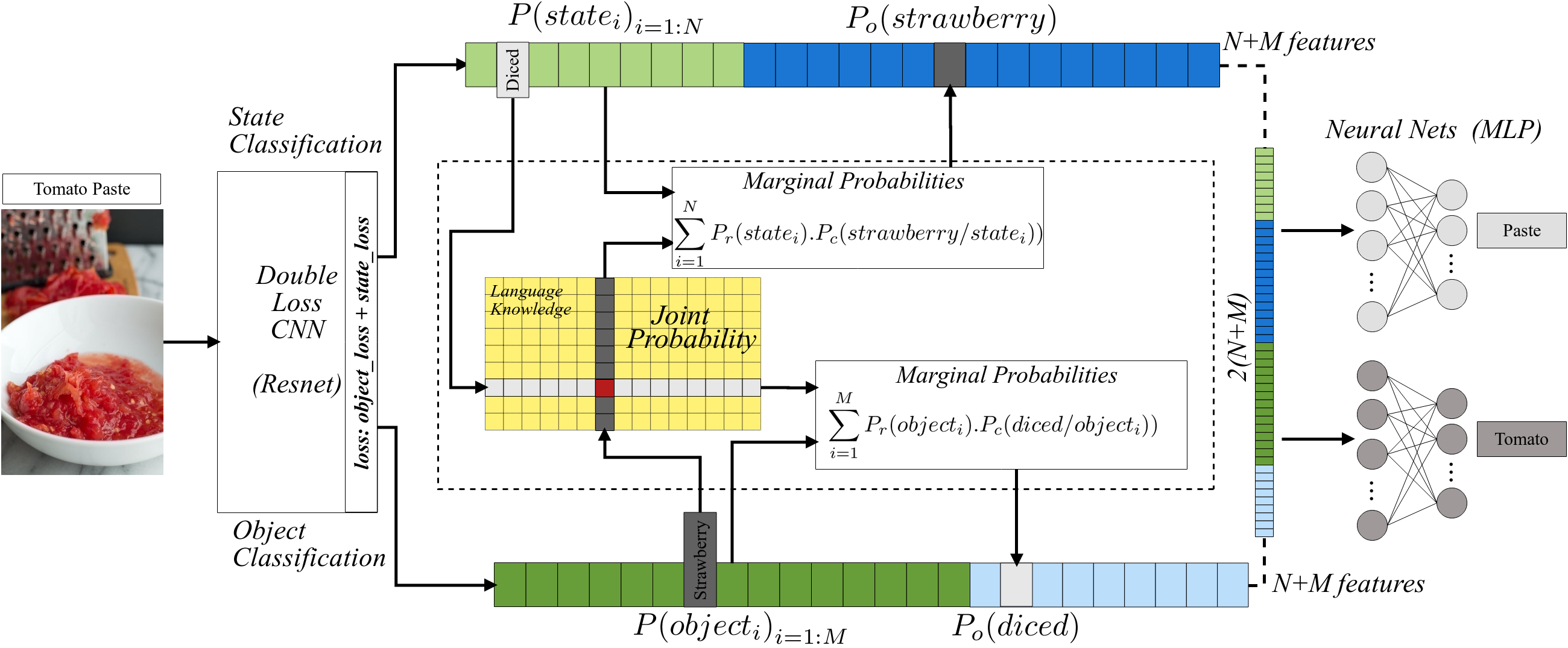}
   \caption{Pipeline for State and Object Classification using Language Knowledge.}
\label{fig:joint_state_and_object}
\end{figure*}

In this paper, we present a pipeline consisting of a deep convolutional network and a language knowledge graph inference strategy for joint state and object classification as shown in Figure \ref{fig:joint_state_and_object}.
The Resnet-50 architecture from \cite{Resnet} is trained with two parallel classification layers for object (e.g. potato) and state (e.g. diced) classification.
Joint confidences for each pair of object and states are computed using the relatedness assertion in Concept-Net.
The object and state marginal probabilities are computed using the confidences from the deep network and the joint confidences derived from Concept-Net.
The outputs from Resnet-50 are concatenated with the inferred marginal probabilities which are then fed to two multi-layer perceptron (MLP) networks. 
The MLPs are trained and the whole pipeline is evaluated over a dataset of cooking objects.
A selector gate is trained to predict whether the CNN model will predict correctly or incorrectly given an input image and is incorporated in the model for prediction improvement.
Our work has two main contributions:
\begin{itemize}
  \item A new pipeline for joint state and object classification which incorporates language knowledge to help with state and object predictions.
  \item A selector gate that improves classification accuracy by utilizing the input and output of a trained classifier.
  \end{itemize}

The rest of the paper is organized as follows. Section \ref{sec:related} introduces the related work in state classification.
Section \ref{sec:method} introduces the methodology including the language knowledge used for state and object classification.
Section \ref{sec:experiments} discusses experiments and results and Section \ref{sec:conclusion} concludes the paper.

\section{Related Work}
\label{sec:related}

Object classification and detection are very popular areas of research \cite{Resnet,Alexnet,VGG,Googlenet}, but state classification from a single image requires more investigation.
Some work explicitly address the states \cite{Ahmad_Paper},
and some perform state identification implicitly \cite{added2_densecap}.
In \cite{Review_ActionRecognition1}, action attributes and parts are used as states of an action for action classification.
High level image attributes have also been incorporated in CNNs and LSTMs to provide descriptions for an image \cite{Review_ImageCaption1}.
In \cite{Ahmad_Paper}, a dataset of states for cooking objects was introduced and the problem of state classification in cooking related images was addressed.
We use this dataset in our work.
In \cite{States_Transforms}, states and state transformations between objects including cooking objects are analyzed on a collection of images.

In \cite{added4_objectstates}, states of objects and state-modifying actions are jointly detected using a discriminative clustering cost. 
In \cite{added1_multi_task_cnn}, a multi-task CNN is proposed for binary attribute prediction. 
Each binary attribute can be considered as a state in our context.
In \cite{Review_ImageCaption4}, a deep convolutional and recurrent framework is presented for providing multiple object labels from a single image. 
This work is similar to our work in the aspect that it provides multiple labels for a single image.
Facial expression can also be considered as a state of the face. In \cite{Identity_Aware}, a multi-loss architecture is proposed to capture both identity and expression associated features for face expression classification.
In \cite{Structural_RNN}, a framework is proposed that simultaneously models multiple concepts or states of position in a sequence using an RNN and a spatio-temporal graph.

Knowledge representations have been effectively used in  combination with classification approaches. 
In \cite{IEEE_TMM}, a video understanding framework was proposed that deploys a deep convolutional network together with a knowledge representation. 
Knowledge representations can also be incorporated in robotics applications \cite{David_Survey} and to aid robots in manipulation decisions for cooking actions \cite{FOON,Foon_extended}.
Concept-Net has also been employed for object detection. 
In \cite{concept_net1}, semantic consistency is seeked by combining information from a knowledge graph such as Concept-Net and any object detection algorithm. 

\section{The Pipeline}
\label{sec:method}

We propose a pipeline for joint state and object identification.
The pipeline includes a convolutional neural network, a language model and two MLP networks as shown in Figure \ref{fig:joint_state_and_object}.
We apply a selector gate on the pipeline outputs to improve results as depicted in Figure \ref{fig:refinement}.

\subsection{Stage 1: Double Loss Convolutional Network}
\label{sec:joint_state_and_object}

In the first stage of the pipeline, we use the Resnet architecture with two outputs- one for state and one for object classification.
The two applications use the same weights apart from the last layer. 
The loss applied for object and state classification are defined separately and trained simultaneously. 
The network outputs two different sets of confidences via the soft-max layer, one for the state classes, \( [{P(state_i)}_{i=1:N_{states}}] \), and another for the object classes, \( [{P(object_i)}_{i=1:N_{objects}}] \), as shown in Figure \ref{fig:joint_state_and_object}.
The notations \( N_{states} \) and \( N_{objects} \) are the number of states and objects respectively.
The soft-max confidences are the first set of probabilities we obtain for object and state classification.
We name them as prior probabilities of each object (or state) occurring in the image.

\subsection{Stage 2: Language Knowledge based Features}
\label{sec:Language_features}

In natural language processing, documents, sentences, and words are processed to extract meanings, relationships and word embeddings.
In this paper we will use the \textit{\textbf{Concept-Net}}, which is more powerful than the widely used Word2vec \cite{Word2Vec}, and the \textit{\textbf{Google N-gram Viewer}} to quantify word relations.

Concept-Net is a language knowledge graph that includes words and phrases as nodes and natural language relationships between the nodes as edges \cite{Conceptnet}.
Concept-Net defines and implements a class of language- and source-independent relations between words and phrases including \textit{IsA}, \textit{UsedFor}, and \textit{CapableOf} and also associates weights with every relationship.
Weights of relations are calculated based on an aggregation of weights from various sources.
We use the weights from the \textit{RelatedTo} relation (or assertion) of the Conceptnet API to quantify the relationship of a specific state (e.g. sliced) with a specific object (e.g. bread).

In natural language processing, an N-gram is a sequence of N items (e.g. words) in a bed of various documents called a corpus \cite{NLP_source}.
The frequency of two or multiple words happening together (N-grams), can be representative of how related 
they are. 
The Google N-gram Viewer is a Google based search engine that shows the frequency of any N words occurring consecutively in Google's text sources \cite{google-ngram}. 
We use the frequencies extracted from the Google N-gram Viewer to represent the relationship between states and objects. 

\subsubsection{Feature Extraction}
\label{sec:feature_extraction}

The correct identification of objects is associated with the correct identification of states and vice versa.
We use the Concept-Net and the Google N-gram Viewer to quantify the relationship between the states and objects in the dataset.
We first define a set of words associated with each object and a set of words associated with each state. 
For instance, for the object \(potato\) we define the set \( \{potato, potatoes\} \) and we define \( \{creamy,paste,mashed,mash,softened,whipped\} \) as the set representing the state \(creamy\).
To calculate the joint probability of an object (e.g. potato) and a state (e.g. creamy), every pair of object and state from the two sets is looked up in Concept-Net or the N-gram Google Viewer to derive a relatedness value.
The maximum and the mean values for each pair are recorded (e.g. potato-creamy). 
The confidences are normalized so that the sum of all probabilities of a state over various objects and the sum of all probabilities of an object with different states each sum up to 1.

We calculate the marginal probabilities for each object assuming the state prior probabilities and joint (conditional) probabilities (\( [{P(object/state_i)}_{i=1:N_{states}}] \)) derived from the language knowledge source (e.g. Concept-net or Google N-gram Viewer).
We conversely compute the marginal probabilities for the states using joint (conditional) probabilities (\( [{P(state/object_i)}_{i=1:N_{objects}}] \)).
The relations for marginal probabilities for each object \( {P(object)} \), and state, \( {P(state)} \), is given in
(\ref{eq:object_probability}), and (\ref{eq:state_probability}) respectively.
\begin{equation}
\label{eq:object_probability}
P_o(object_j) = {\sum_{i=1}^{N_{states}} P_r(state_i).P_c(object_j/state_i))}
\end{equation}
\begin{equation}
\label{eq:state_probability}
P_o(state_j) = {\sum_{i=1}^{N_{objects}} P_r(object_i).P_c(state_j/object_i)}
\end{equation}

In (\ref{eq:object_probability}), and (\ref{eq:state_probability}), \( P_c \) is the conditional probability of an object in respect to a state or vice versa which is derived from the language knowledge, \( P_r \) is the output confidence from the Resnet, and \( P_o \) is the marginal probability. 

\subsection{Stage 3: Neural Network Predictions}
\label{sec:Neural_Nets}

The marginal and prior probabilities are concatenated together to create a feature vector of size \(2 \times N_{objects} \) and \(2 \times N_{states} \) for objects and states respectively.
The concatenated object and state features are merged together to create a final feature vector with size \(V_{final} = 2 \times (N_{states}+N_{objects}) \). 
The feature vector \(V_{final}\) is given as input to two separate MLP networks for object and state classification respectively as shown in Figure \ref{fig:joint_state_and_object}.
A three layer MLP is selected using the validation set as the finalized architecture of the networks.

\subsection{Stage 4: Model Refinement}
\label{sec:refinement}
\begin{figure}
   \centering
   \includegraphics[width=8cm]{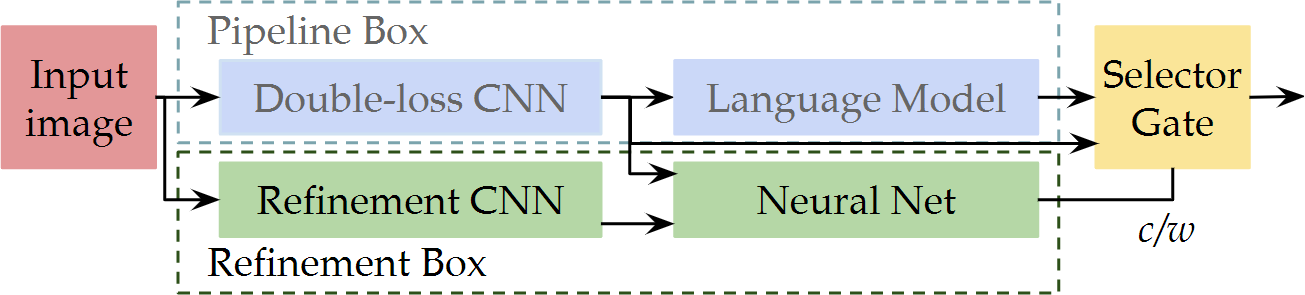}
   \caption{Pipeline Refinement.}
\label{fig:refinement}
\end{figure}

The pipeline converts correct predictions into incorrect predictions in some cases.
To reduce these conversions, a refinement procedure is proposed that starts training after the double loss CNN has finished training.
The refinement model is trained to predict the probability of an image being classified correctly by the pipeline.
The refinement model contains a Resnet-based CNN which returns two outputs (classes) for a given input image; one output represents an image being classified correctly and the other represents the image being classified incorrectly.
The two outputs are associated with two confidences.
The two confidences are concatenated with the ouput probabilities from the double loss CNN from the pipeline.
Two separate neural networks are trained for correct/incorrect object (and state) probability predictions using the concatenated feature vectors.
The outputs from the MLP are used as a selector for a gate selector block.
A value of one for the selector output, means that the initial prediction is correct and the object (or state) confidences from the double loss CNN are used for predictions. 
A value of zero for the selector output, means that the the probabilities after language knowledge incorporation should be used for predictions.
The refinement model is depicted in Figure \ref{fig:refinement}.

\section{Experiments and Results}
\label{sec:experiments}

\subsection{Dataset}
\label{sec:dataset}

\begin{figure}
   \centering
   \includegraphics[width=8cm]{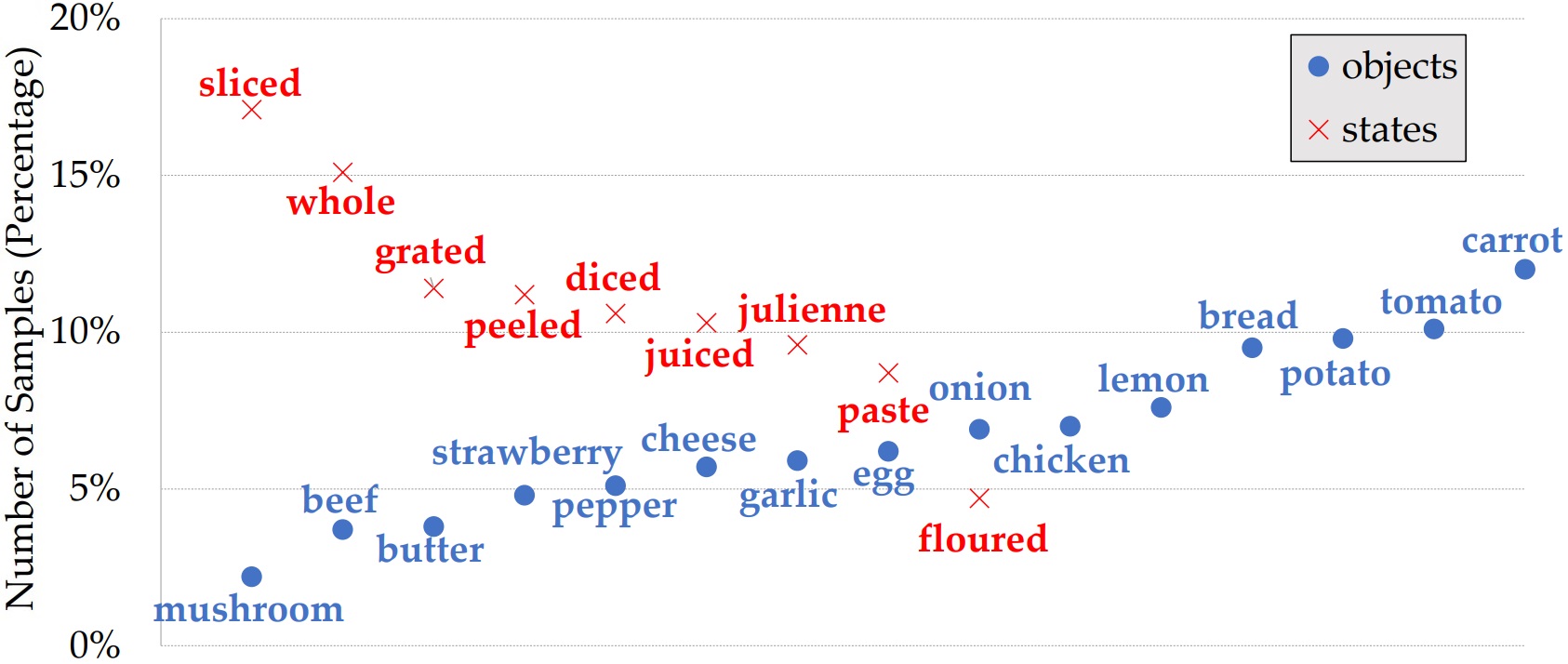}
   \caption{States and Objects statistics in the dataset.}
\label{fig:states}
\end{figure}

We used the state classification dataset from \cite{Ahmad_Paper}.
It consists of images from 15 cooking objects and 11 state classes as shown in Figure \ref{fig:states}.
For our experiments, we removed the states \textit{mixed} and \textit{other} that are not associated with any specific type of object.
When training, we use data augmentation to balance the classes and compute the accuracy as average class accuracy.
We annotate the dataset with object labels.
The total number of images in the dataset is around 9.5K. -- 70\% train, 15\% validation, and 15\% test set. 
The dataset includes an online challenge page with the best state classification results ranked from best to worst\footnote{\href{http://rpal.cse.usf.edu/datasets\_cooking\_state\_recognition.html}{http://rpal.cse.usf.edu/datasets\_cooking\_state\_recognition.html}}.
We used the statistical information from the knowledge representation in \cite{FOON} to derive the most frequent objects and states represented in cooking events. 
States were analyzed hierarchically, and the main states associated with the most frequent objects were derived \cite{Ahmad_Paper}. 

\subsection{Results}
\label{sec:exps}
We implemented the Resnet model in Tensorflow and initialized with pre-trained weights from Imagenet.
The single classifier layer was removed and a double classifier layer was added for states and objects.
We trained the model for 15 iterations and with an initial learning rate of 0.01.
Only weights from the last block of Resnet were trained and the rest were kept frozen.
The relatedness values of objects and states were downloaded from the Concept-Net (or Google N-gram Viewer) Web APIs using the Python Request library and the normalized versions of the relatedness values were recorded as joint probabilities.
The final features were then computed and then given to MLPs as mentioned in Subsection \ref{sec:Neural_Nets}. 

We compared the pipeline with other methods and report the results in Table \ref{table_results}.
We compared the pipeline with the raw initial confidences, the linear combination of the initial confidence and the marginal probabilities from Concept-Net, and an SVM-based version of the pipeline.
The results show that all methods containing a language knowledge outperform the Resnet network as shown in Table \ref{table_results}.
The neural network based method that uses features from the Resnet output and the Concept-Net features outperforms all other methods.
Results in Table \ref{table_results} show that self-correction using the refinement model improves the results even further.

\begin{table} [!ht]
\centering
\caption{States and object classification accuracy on the test set with and without using Concept-Net (Concept-Net as CN, Google N-gram as GN).}
\label{table_results}
 \begin{tabular}{|c |c |c |c |}
 \hline
\textbf{Model} & \textbf{States} & \textbf{Objects}\\ [0pt]
 \hline
 Resnet & 79.4\% & 74.1\% \\ [0pt]
 \hline
 (Resnet,CN) + SVM & 79.7\% & 74.2\% \\ [0pt]
 \hline
 (Resnet,GN) + MLP & 80.1\% & 74.2\% \\ [0pt]
 \hline
 (Resnet,CN) + MLP & 80.4\% & 74.3\% \\ [0pt]
 \hline
 (Resnet,CN) + MLP + Refinement & \textbf{80.9}\% & \textbf{75}\% \\ [0pt]
 \hline
\end{tabular}
\end{table}

\begin{figure} [!h]
   \centering
   \includegraphics[width=8cm]{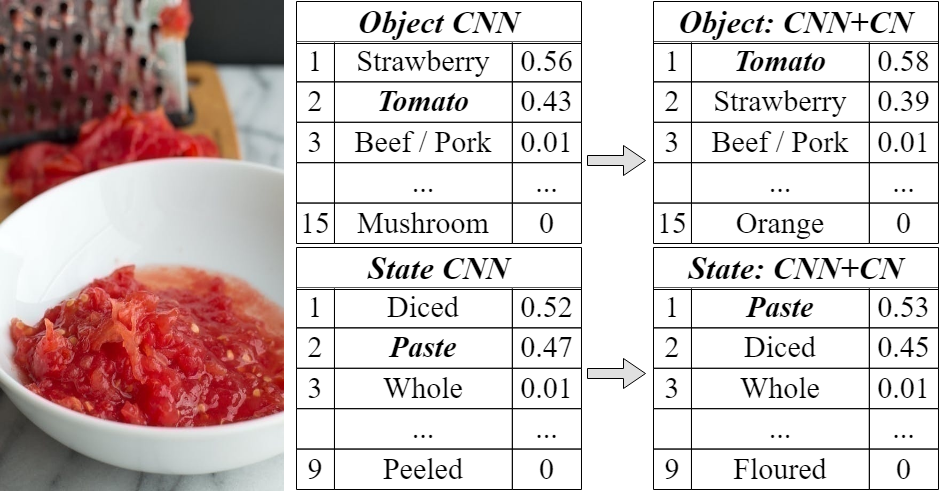}
   \caption{Objects and states CNN probabilities vs Concept-Net (CN) probabilities. Probability of diced strawberry is lower than tomato paste when CN is used.}
\label{fig:CN_vs_DNN}
\end{figure}

Figure \ref{fig:CN_vs_DNN} shows an instance of an incorrect result (diced strawberry) converting to a correct result (tomato paste) when using Concept-Net.
Concept-Net can make mistakes.
For example grated butter has a high relatedness confidence in the Concept-Net graph although in the real world it is unlikely to see grated butter often.
Therefore, it is easy to flip a correct \textit{creamy butter} to an incorrect \textit{grated butter}.
The refinement model has the ability to prevent some of these cases.

%\subsection{Discussion}
%\label{sec:discussion}
%In robotics applications, knowledge about the state or size of a cooking object is of importance in various aspects such as robotic manipulation  \cite{Ahmad_Paper}, visual recognition \cite{States_Transforms}, knowledge representation \cite{FOON}, and robotic feeding \cite{robotic_feeding}.
%In \cite{FOON}, a knowledge representation of cooking objects, their affordances, and events related to them was proposed. 

\section{Conclusion}
\label{sec:conclusion}

The states of a cooking object are valuable information for a robot chef when performing cooking events and are closely related with the object itself. 
This paper presented a deep neural network with two joint losses for object and state classification. 
A language knowledge graph was deployed on top of confidences from a double loss CNN for extracting language based confidences. 
A MLP-based classifier was trained using the combination of confidences from both stages.
Experiments on a state classification dataset consisting of cooking objects showed that using a language knowledge together with the confidences from the deep network improved both object and state classification performance.

\bibliographystyle{IEEEbib}
\bibliography{strings,refs}

\begin{thebibliography}{10}

\bibitem{Resnet}
K.~He, X.~Zhang, S.~Ren, and J.~Sun,
\newblock ``Deep residual learning for image recognition,''
\newblock {\em {CVPR}}, pp. 770--778, 2016.

\bibitem{Review_ActionRecognition1}
B.~Yao, X.~Jiang, A.~Khosla, A.~L. Lin, L.~Guibas, and L.~Fei-Fei,
\newblock ``Human action recognition by learning bases of action attributes and
  parts,''
\newblock {\em ICCV}, pp. 1331--1338, Nov 2011.

\bibitem{Review_ImageCaption1}
T.~Yao, Y.~Pan, Y.~Li, Z.~Qiu, and T.~Mei,
\newblock ``Boosting image captioning with attributes,''
\newblock {\em ICCV}, vol. 00, pp. 4904--4912, Oct. 2018.

\bibitem{added4_objectstates}
J.~B. Alayrac, J.~Sivic, I.~Laptev, and S.~Lacoste-Julien,
\newblock ``Joint discovery of object states and manipulation actions,''
\newblock {\em ICCV}, 2017.

\bibitem{added1_multi_task_cnn}
A.~H. Abdulnabi, G.~Wang, J.~Lu, and K.~Jia,
\newblock ``Multi-task cnn model for attribute prediction,''
\newblock {\em IEEE Transactions on Multimedia}, vol. 17, pp. 1949--1959, 2015.

\bibitem{Ahmad_Paper}
A.~B. {Jelodar}, M.~S. {Salekin}, and Y.~{Sun},
\newblock ``Identifying object states in cooking-related images,''
\newblock {\em arXiv preprint arXiv:1805.06956}, May 2018.

\bibitem{concept_net1}
Y.~Fang, K.~Kuan, J.~Lin, C.~Tan, and V.~Chandrasekhar,
\newblock ``Object detection meets knowledge graphs,''
\newblock {\em IJCAI-17}, pp. 1661--1667, 2017.

\bibitem{Word2Vec}
T.~Mikolov, K.~Chen, G.~Corrado, and J.~Dean,
\newblock ``Efficient estimation of word representations in vector space,''
\newblock {\em ICLR}, May 2013.

\bibitem{Alexnet}
A.~Krizhevsky, I.~Sutskever, and G.~E. Hinton,
\newblock ``Imagenet classification with deep convolutional neural networks,''
\newblock {\em NIPS}, vol. 1, pp. 1097--1105, 2012.

\bibitem{VGG}
K.~Simonyan and A.~Zisserman,
\newblock ``Very deep convolutional networks for large-scale image
  recognition,''
\newblock {\em ICLR}, 2015.

\bibitem{Googlenet}
C.~Szegedy, W.~Liu, Y.~Jia, P.~Sermanet, S.~Reed, D.~Anguelov, D.~Erhan,
  V.~Vanhoucke, and A.~Rabinovich,
\newblock ``Going deeper with convolutions,''
\newblock {\em CVPR}, 2015.

\bibitem{added2_densecap}
J.~Johnson, A.~Karpathy, and L.~Fei-Fei,
\newblock ``Densecap: Fully convolutional localization networks for dense
  captioning,''
\newblock {\em CVPR}, 2016.

\bibitem{States_Transforms}
P.~Isola, J.~J. Lim, and E.~H. Adelson,
\newblock ``Discovering states and transformations in image collections,''
\newblock {\em CVPR}, 2015.

\bibitem{Review_ImageCaption4}
J.~Wang, Y.~Yang, J.~Mao, Z.~Huang, C.~Huang, and W.~Xu,
\newblock ``Cnn-rnn: A unified framework for multi-label image
  classification,''
\newblock {\em CVPR}, 2016.

\bibitem{Identity_Aware}
Z.~Meng, P.~Liu, J.~Cai, S.~Han, and Y.~Tong,
\newblock ``Identity-aware convolutional neural network for facial expression
  recognition,''
\newblock {\em FG 2017}, pp. 558--565, May 2017.

\bibitem{Structural_RNN}
A.~Jain, A.~R. Zamir, S.~Savarese, and A.~Saxena,
\newblock ``Structural-rnn: Deep learning on spatio-temporal graphs,''
\newblock {\em CVPR}, pp. 5308--5317, June 2016.

\bibitem{IEEE_TMM}
A.~B. Jelodar, D.~Paulius, and Y.~Sun,
\newblock ``Long activity video understanding using functional object-oriented
  network,''
\newblock {\em IEEE Transactions on Multimedia}, pp. 1--12, 2018.

\bibitem{David_Survey}
D.~Paulius and Y.~Sun,
\newblock ``A survey of knowledge representation in service robotics,''
\newblock {\em Robotics and Autonomous Systems}, 2019.

\bibitem{FOON}
D.~Paulius, Y.~Huang, R.~Milton, W.~D. Buchanan, J.~Sam, and Y.~Sun,
\newblock ``Functional object-oriented network for manipulation learning.,''
\newblock {\em IROS}, pp. 2655--2662, 2016.

\bibitem{Foon_extended}
D.~Paulius, A.~B. Jelodar, and Y.~Sun,
\newblock ``Functional object-oriented network: Construction \& expansion,''
\newblock {\em ICRA}, pp. 1--7, May 2018.

\bibitem{Conceptnet}
R.~Speer, J.~Chin, and C.~Havasi,
\newblock ``Conceptnet 5.5: An open multilingual graph of general knowledge,''
\newblock {\em AAAI}, 2016.

\bibitem{NLP_source}
C.~D. Manning and H.~Shutze,
\newblock ``Foundations of statistical natural language processing,''
\newblock in {\em The MIT Press}, 1999.

\bibitem{google-ngram}
Google,
\newblock ``Google ngram viewer,'' http://books.google.com/ngrams/datasets,
  2012.

\end{thebibliography}

\end{document}